\documentclass[letterpaper, 10 pt, conference]{ieeeconf}  

\IEEEoverridecommandlockouts                              

\usepackage{cite}
\usepackage{graphicx}
\usepackage[ruled, vlined]{algorithm2e}
\usepackage{url}
\usepackage{amsmath} 
\usepackage{amssymb}  
\usepackage[pagebackref=true,breaklinks=true,colorlinks,bookmarks=false]{hyperref}
\usepackage[capitalize]{cleveref}
\usepackage{siunitx}
\usepackage{booktabs}
\usepackage{multirow}
\usepackage{multicol}
\usepackage{float}
\usepackage{pifont}

\usepackage{times} 

\newcommand{\cc}{\color[rgb]{0,0.6,0.3}\ding{52}}
\newcommand{\ccc}{\color[rgb]{1.0,0.8,0.0}\ding{52}}
\newcommand{\xx}{\color[rgb]{0.6,0,0}{\ding{55}}}

\usepackage{colortbl}
\definecolor{lightgray}{gray}{0.9} 

\title{\LARGE \bf
Lightweight Event-based Optical Flow Estimation\\
via Iterative Deblurring
}

\author{Yilun Wu$^{1}$, Federico Paredes-Vallés$^{1}$ and Guido C. H. E. de Croon$^{1}$
\thanks{$^{1}$All authors are with the Micro Air Vehicle Laboratory, Faculty of
Aerospace Engineering, Delft University of Technology, Delft, The Netherlands. Correspondences:
        {\tt\small Y.Wu-9@tudelft.nl}}%
}

\begin{document}

\maketitle
\thispagestyle{empty}
\pagestyle{empty}

\begin{abstract}

Inspired by frame-based methods, state-of-the-art event-based optical flow networks rely on the explicit construction of correlation volumes, which are expensive to compute and store, rendering them unsuitable for robotic applications with limited compute and energy budget. Moreover, correlation volumes scale poorly with resolution, prohibiting them from estimating high-resolution flow. We observe that the spatiotemporally continuous traces of events provide a natural search direction for seeking pixel correspondences, obviating the need to rely on gradients of explicit correlation volumes as such search directions. We introduce \textit{IDNet} (Iterative Deblurring Network), a lightweight yet high-performing event-based optical flow network directly estimating flow from event traces without using correlation volumes. We further propose two iterative update schemes: ``\textit{ID}" which iterates over the same batch of events, and ``\textit{TID}" which iterates over time with streaming events in an online fashion. Our top-performing \textit{ID} model sets a new state of the art on DSEC benchmark. Meanwhile, the base \textit{ID} model is competitive with prior arts while using 80\% fewer parameters, consuming 20x less memory footprint and running 40\% faster on the NVidia Jetson Xavier NX. Furthermore, the \textit{TID} model is even more efficient offering an additional 5x faster inference speed and 8 ms ultra-low latency at the cost of only a 9\% performance drop, making it the only model among current literature capable of real-time operation while maintaining decent performance.

Code: \url{https://github.com/tudelft/idnet}.

\end{abstract}

\section{Introduction}
\label{sec:intro}

\begin{figure}[t]
  \centering
    \includegraphics[width=0.89\linewidth]{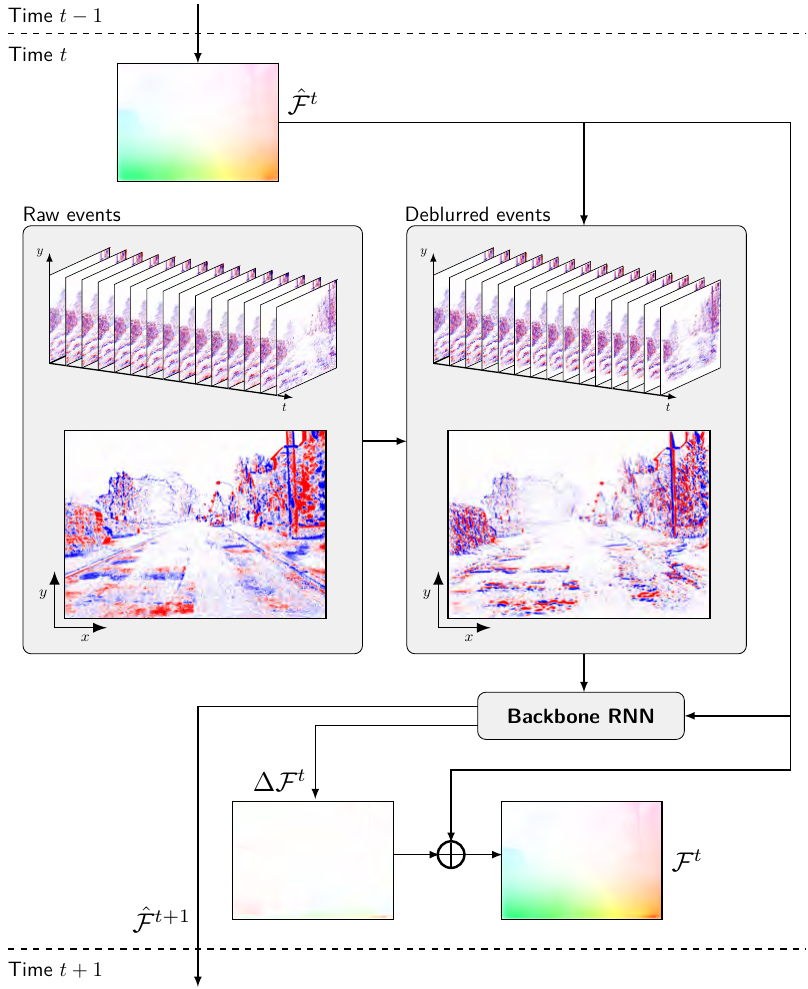}
  \caption{Illustration of the \textit{IDNet} pipeline for temporal iterative deblurring (i.e. \textit{TID} scheme). Raw events are first deblurred according to the initial coarse optical flow estimate $\hat{\mathcal{F}}^t$ before being processed by the backbone RNN. The RNN extracts the residual motion from the deblurred events and outputs the residual flow $\Delta \mathcal{F}^t$ which is added to the initial estimate $\hat{\mathcal{F}}^t$ to arrive at the final flow estimation $\mathcal{F}^t$. The RNN additionally proposes a coarse estimate $\hat{\mathcal{F}}^{t+1}$ for the next timestep under continuous operation.}
  \label{fig:cover}
\end{figure}

Optical flow estimation, i.e.\ estimating pixel motion over time on the image plane, is both a central and challenging task in computer vision. As it encodes a primitive form of motion information, optical flow underpins many robotic navigation applications \cite{ flownavigation,qin2017vins, nanoflownet}. Compared to frame-based cameras, event cameras capture asynchronous brightness changes in continuous time, offering high dynamic range measurements with minimal motion blur at high speeds and low lighting conditions while only consuming milliwatts of power \cite{event-vision-survey}. All these advantages make it the ideal sensor candidate for resource-constrained agile robots such as micro aerial/ground vehicles (MAVs/MGVs) \cite{howfast, event-quadrupedal}. 

Learning-based methods have made marked progress in optical flow estimation by incorporating more apt inductive biases \cite{pwc, flownet, raft}. Notably, the dense all-pair correlation volumes introduced in \cite{raft} effectively estimate high-quality flow for large motions and have been widely adopted in later research \cite{gma, flowformer, gmflow}. Recent methods for event-based optical flow \cite{eraft, tma, blinkflow} emulate these approaches, treating consecutive event frames as discrete images to determine flow.

However, three major drawbacks of correlation volumes rise when applied to event data. First, constructing them requires accumulation of events which causes high latency. Second, computing and storing the correlation volume is expensive, restricting deployment on memory/compute-limited systems. Finally, as noted in \cite{scv, 1dcorr}, the correlation volume scales poorly to higher input resolutions, limiting the algorithm's ability to process higher resolution events or deliver fine details in optical flow estimates.

We observe that, compared to estimating flow from two images, event-based optical flow benefits from the continuous recording of motion over time and space. This allows tracing the continuous trajectory to estimate flow, whereas frame-based methods have to search for such ``trajectory" by estimating gradients in the correlation volume. In other words, while correlation volumes are instrumental for methods operating over frames to propose an update direction, such direction is readily evident in the form of blur (i.e. the continuous spatiotemporal trajectory) in the raw events.

From this observation, we propose \textit{IDNet} (Iterative Deblurring Network), an event-based algorithm for iterative optical flow estimation without correlation volumes. At its core, \textit{IDNet} processes event bins sequentially via a backbone RNN where flow is estimated from the traces (i.e. blur) of events. Additionally, we adopt an iterative update scheme to improve estimation quality under rapid motion, employing motion compensation (a.k.a.\ deblurring) \cite{contrastmax}. Each iteration deblurs events based on prior flow estimates, then processes them to refine the flow, akin to frame-based warping techniques.

We propose two iterative update schemes: \textit{ID} (iterative deblurring), illustrated in 
\cref{fig:id-pipeline}, which iterates over the same batch of events to achieve the best performance, and \textit{TID} (temporal iterative deblurring), shown in \cref{fig:cover}, which iterates over events in time for drastically faster processing.

On public benchmarks, our methods achieve comparable results with state-of-the-art methods that use correlation volumes, while using much fewer parameters and memory. Without correlation volumes, our method can estimate optical flow from higher-resolution feature maps, resulting in significant improvements over prior art. Additionally, our \textit{TID} scheme is highly efficient, reaching close-to-state-of-the-art performance while incurring minimal latency.

\section{Related Works}
\label{sec: relatedworks}

\subsection{Event-based Optical Flow}
Traditionally, optical flow estimation is performed from image frames with algorithms ranging from classical optimization techniques such as \cite{schunck, LKT} to modern learning-based methods \cite{flownet, pwc, raft, flowformer, gma, gmflow} which combine the powerful expressivity of deep neural networks with inductive biases such as correlation matching and coarse-to-fine estimation. 

Despite the maturity of such solutions, images would suffer from quality degradation such as motion blur and insufficient exposure. Event cameras, on the other hand, are not plagued by such issues \cite{event-vision-survey} and hence are more suited for use in these challenging conditions.

Early event-based optical flow algorithms such as \cite{BENOSMAN201232} adapt frame-based methods such as KLT \cite{LKT} to the event-based domain or utilize hand-crafted heuristics to fit event data \cite{benosman2014, event-lifetime}. The approach in \cite{sofie} jointly optimizes image brightness and  flow by exploiting brightness constancy. Later, a framework to recover optical flow from events by maximizing the contrast of motion-compensated events is introduced in \cite{contrastmax, secret}. Other than directly optimizing the objective, this principle is also applied as a learning signal in self-supervised learning methods such as \cite{neurips_snn, event-basics, 9341224, ev-flownet}.

The availability of event simulators \cite{vid2e, v2e, blinkflow} and large-scale datasets \cite{dsec, mvsec, m3ed} enable learning-based methods to achieve superior performance over model-based ones. EV-Flownet \cite{ev-flownet} introduces an event representation which split events proportionally to the nearby temporal bins and a U-Net architecture for processing the event representation. ECN \cite{9341224} jointly estimates optical flow, depth and egomotion. Both methods construct multi-level feature pyramids. E-RAFT \cite{eraft}, an event-based version of RAFT \cite{raft} is the first to introduce correlation volumes into the event domain. The method computes the 4D all-pair correlation volume between two neighboring event representations in time, which is then iteratively processed to arrive at the final optical flow estimate. TMA \cite{tma} improves upon \cite{eraft} by constructing multiple correlation volumes at a finer temporal scale.

Although such methods \cite{tma, eraft} achieve high performance, the explicit computation and storage of correlation volumes, as mentioned in \cite{scv, 1dcorr}, has poor time and storage complexity which scales badly with input resolution. For events with resolution H by W, the compute and storage complexity of such correlation volume are $\mathcal{O}({(\text{H} \times \text{W})}^2)$. This storage requirement eliminates the possibility of its deployment on memory-limited compute platforms and prohibits the algorithm from scaling up to larger input resolutions.


\subsection{Iterative Refinement for Optical Flow}
The idea of iterative refinement for optical flow can be traced back to the early works of iterative KLT tracking \cite{LKT} where a Taylor series expansion is applied to linearize the problem and iteratively solve for the residual error. Learning-based methods incorporate this principle as well. Most works \cite{spynet, pwc, liteflownet} perform coarse-to-fine refinement along the feature pyramid. In addition, other works such as \cite{IRR, flownet2, devon, raft} stack multiple modules in series to iteratively estimate the residual flow. While some warps images \cite{flownet2}, other methods such as \cite{IRR, raft, devon} resample correlation volumes. 

Recently, iterative refinement for optical flow has also been applied to event-based data. Methods such as \cite{scflow} and \cite{event-pwc} use pyramid scale to perform coarse-to-fine estimation similar to PWC-Net \cite{pwc} but do not perform flow refinement through iterations. The update scheme from \cite{IRR} is adopted by \cite{ding2021spatio}, which processes the correlation volume between warped event slices through time to arrive at a residual flow estimate. E-RAFT \cite{eraft} instead directly processes the correlation volume through an RNN and extracts the final flow estimate from the final state of the RNN. All aforementioned methods rely on the explicit construction of correlation volumes as a basis for their refinement schemes, in part to adapt existing successful architectures in the frame-based domain to event data. 


\subsection{Continuous Operation}
Since optical flow is temporally highly correlated as the motion in a real-world environment is mostly smooth and continuous, it makes sense to leverage the prior estimate from the past to better predict the present optical flow. This temporal recurrency is introduced in \cite{continual_flow} and used in RAFT \cite{raft} as a ``warm-starting" strategy when applied to video data. RAFT \cite{raft}, as well as E-RAFT \cite{eraft}, directly transports the previous flow field to the current timestep by the flow itself under the assumption of constant linear motion. 

\section{Method}
\label{sec:method}

\subsection{Event Representation}
As in \cite{eraft, Zhu_2019}, we utilize the discretized event voxel grid as the event representation. Specifically, we create a representation $\mathcal{E} \in \textrm{R}^{\textrm{B}\times \textrm{H} \times \textrm{W}}$ with $\textrm{B}$ bins out of the event stream $\{e_i = (t_i, x_i, y_i, p_i)\}$ of interest as follows:
\begin{align}
t_i^\ast&=(B-1)(t_i-t_\text{begin})/(t_\text{end}-t_\text{begin}) \\
\mathcal{E}(b, x, y) &= \sum_{i|x_i=x, y_i=y} p_i \max(0, 1-|b-t_i^\ast|) 
\end{align}

where $b$ is the integer bin index and $\textrm{H}, \textrm{W}$ correspond to the height and width resolution of the event stream respectively.

\begin{figure*}[ht]
  \centering
  \begin{minipage}[c]{0.73\linewidth}
    \includegraphics[width=\linewidth]{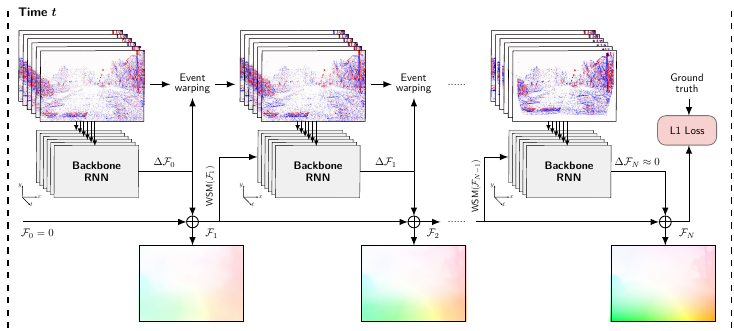}
    \end{minipage}
    \hfill
    \begin{minipage}[c]{0.26\linewidth}
    \includegraphics[width=\linewidth]{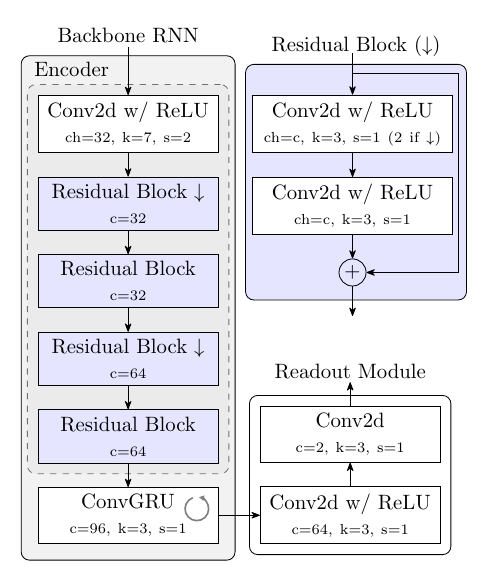}
    \end{minipage}
    \caption{Overall pipeline of \textit{IDNet} with iterative deblurring scheme (i.e. \textit{ID} scheme). Starting with a zero flow, each iteration deblurs events using prior flow. The deblurred event bins are fed into the RNN sequentially one bin at a time. A residual flow is estimated and used for deblurring in the next iteration. The final flow accumulates all residual flows throughout iterations. An L1 loss is applied between the final flow estimate and ground truth. The detailed network structure is shown on the right. The parameters ch, k, and s of the Conv2d layer refer to the output channel count, kernel size, and stride.}
    \label{fig:id-pipeline}
    \vspace{-0.5cm}
\end{figure*}
\subsection{Motion Compensation}
We utilize the principle of motion compensation \cite{contrastmax} as a core step in our processing. Specifically, we use the flow estimate $\mathcal{F}$ to deblur the events to a reference time $t_\text{ref}$ by changing its pixel coordinate $\mathbf{x}_i = (x_i, y_i)$ as follows:
\begin{equation}
    \mathbf{x'}_i = \mathbf{x}_i + (t_{\text{ref}} - t_i)\mathcal{F}(x_i)
    \label{eq: deblur}
\end{equation}
Assuming brightness constancy and linear motion, the ideal flow should neutralize motion in events, resulting in a static representation without any motion. The effect of such motion compensation is depicted in \cref{fig:cover}. By projecting all bins of event representation $\mathcal{E}$ onto the same 2D plane, deblurred events show reduced motion, making moving objects appear static with sharp edges. Refer to the supplementary video for a dynamic illustration.


\subsection{Iterative Deblurring}

The main architecture of \textit{IDNet} under iterative deblurring (\textit{ID} scheme) is shown in \cref{fig:id-pipeline}. The backbone of our proposed \textit{IDNet} is a recurrent neural network which processes the input event bins sequentially one bin at a time. An optical flow field is read out from the state of the RNN once all bins have been passed through it. In this case, we select a ConvGRU as the choice for the recurrent unit.

Next, we introduce iterative deblurring on top of the sequential processing described above. Our design is inspired by the predictive coding scheme in the visual cortex \cite{predictive-coding}. In this scheme, our perception of the world is constantly updated, with only the error signal, i.e. the difference between actual visual input and our current prediction, being relayed through the hierarchy of visual cortex for refinement. Analogously, we utilize motion compensation as our prediction model, iterating to refine the flow from the residual signal.

\begin{algorithm}[t]
\small
 \DontPrintSemicolon
 \SetKwInOut{Input}{input}\SetKwInOut{Output}{output}
 \SetKwFunction{RNN}{RNN}
 \SetKwFunction{WarmStartModule}{WarmStartModule}
 \SetKwFunction{Deblur}{Deblur}
 \SetKwFunction{ForwardProp}{ForwardProp}
 \SetKwFunction{E}{Encoder}
 \Input{Event bins $\mathcal{E}$ containing events of duration $\text{T}$ \\
 Number of deblurring iterations $N$}
 \Output{Optical Flow Prediction $\mathcal{F}$ for the motion during the time window of events}
 \BlankLine
  $\mathcal{F}_{0} \leftarrow  \mathbf{0}$, $\mathcal{M}_{\text{RNN}, 0} \leftarrow \mathbf{0}$, $\mathcal{E}_{\text{deblur}, 0} \leftarrow \mathcal{E}$\;
  \For{$i \leftarrow 0; i<N; i$++}{
    $\Delta \mathcal{F}_{i} \leftarrow $ \RNN{\E{$\mathcal{E}_{\text{deblur},i}$}; $\mathcal{M}_{\text{RNN}, i}$} \;
    $\mathcal{F}_{i+1} \leftarrow \mathcal{F}_{i} + \Delta \mathcal{F}_{i}$\;
    $\mathcal{M}_{\text{RNN},i+1} \leftarrow$ \WarmStartModule{$\mathcal{F}_{i+1
    }$} \;
    $\mathcal{E}_{\text{deblur},i+1} \leftarrow$ \Deblur{$\mathcal{E}_{\text{deblur},i}; \Delta \mathcal{F}_{i}$}
  }
  $\mathcal{F} \leftarrow \mathcal{F}_N$ \;
  \KwRet{$\mathcal{F}$} \;
 \caption{Iterative deblurring (\textit{ID}).}
 \label{algo:id}
\end{algorithm}

An overview of the algorithm is presented in \cref{algo:id}. Specifically, as shown in \cref{fig:id-pipeline}, we start with a zero-initialized flow estimate and backbone RNN memory ($\mathcal{F}_0 = \mathbf{0}$, $\mathcal{M}_{\text{RNN},0}=0$). After the RNN predicts the flow $\mathcal{F}_1$ from the raw event bins $\mathcal{E}_0$, the event bins are motion-compensated with the optical flow vector $\mathcal{F}_1(x_i)$, resulting in partially deblurred event bins $\mathcal{E}_1$. This starts the iterative refinement process, with diminishing flow refinement over iterations.

To leverage the RNN's recurrency, we use a warm-starting module (WSM) comprising feed-forward convolutional layers to set the memory of our backbone RNN: $\mathcal{M}_{\text{RNN},1}=\text{WSM}(\mathcal{F}_1)$. We then update the previous belief by adding the residual flow $\Delta \mathcal{F}_2$ estimated from $\mathcal{E}_1$ to arrive at the flow estimate at iteration 2: $\mathcal{F}_2 = \mathcal{F}_1 + \Delta \mathcal{F}_2$, from which a new iteration could be started again. 

Weight sharing the RNN for all iterations is motivated by the expectation that the backbone network can extract any residual motion in the event representation, raw or deblurred, rendering the task unchanged across iterations.

\subsection{Iterative Deblurring through Time}
\label{sec: idtt}

\begin{figure}
    \centering
    \includegraphics[width=\linewidth]{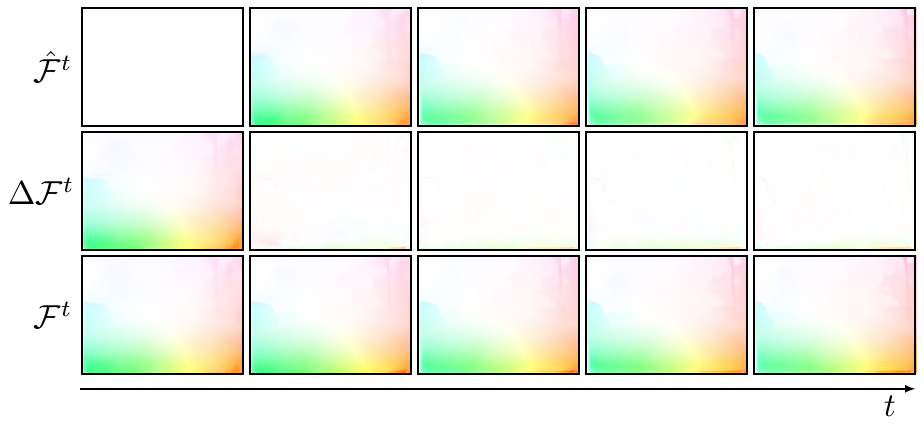}
    \caption{Initial flow $\hat{\mathcal{F}}^t$, residual flow $\Delta \mathcal{F}^t$ and final flow $\mathcal{F}^t$ as time progresses and new event bins keeps coming. A trend of increasing quality in $\mathcal{F}^t$ and lowering magnitude in $\Delta \mathcal{F}^t$ can be observed, implying that the flow is iteratively refined through time.}
    \label{fig:co-flow}
\end{figure}

We also explore another mode of iterative refinement through the temporal scale, as shown in \cref{fig:cover}. Instead of deblurring the same events multiple iterations, we give the network access to a continuous stream of events where at each timestep we deblur the event bins $\mathcal{E}^t$ and pass the deblurred bins $\mathcal{E}_\text{deblur}^{t}$ through the backbone RNN only once. In addition to estimating the optical flow $\mathcal{F}^t$ for the current timestep, we predict an initial flow $\hat{\mathcal{F}}^{t+1}$ for deblurring the event bins $\mathcal{E}^{t+1}$ from the next timestep. This recurrent approach refines flow over time, as shown in \cref{fig:co-flow}. 
\setlength{\abovecaptionskip}{-15pt} 
\setlength{\abovecaptionskip}{15pt}

The key difference between running this pipeline vs simply passing the raw blurry event bins through the backbone network once, is that our pipeline initializes an informative memory from previous estimates and only processes deblurred event bins with smaller motion range. This enables the backbone RNN to improve flow accuracy at a finer scale with limited parameters. 

Operational-wise, we utilize the same motion compensation principle and warm start the RNN memory as before, but introduce an additional module to estimate the initial flow of next timestamp $\hat{\mathcal{F}}^{t+1}$ through the state of the RNN. A pseudo code of the algorithm is laid out in \cref{algo:cid}.
\begin{algorithm}
\small
 \DontPrintSemicolon
 \SetKwInOut{Input}{input}\SetKwInOut{Output}{output}
 \SetKwFunction{RNN}{RNN}
 \SetKwFunction{WarmStartModule}{WarmStartModule}
 \SetKwFunction{Deblur}{Deblur}
 \SetKwFunction{ForwardProp}{ForwardProp}
 \SetKwFunction{E}{Encoder}
 \Input{Event bins $\mathcal{E}^t$ containing events with $\tau\in [t, t+\text{T}]$ \\ Initial flow estimate at  $\hat{\mathcal{F}}^{t}$ at time $t$} 
 \Output{Optical Flow Prediction $\mathcal{F}^t$ at time $t$ for the motion from $\tau \in [t, t+\text{T}]$}
 \BlankLine
  
    $\mathcal{M}_\text{RNN}^{t} \leftarrow$ \WarmStartModule{$\hat{\mathcal{F}}^{t}$} \;
    $\mathcal{E}_\text{deblur}^{t} \leftarrow$ \Deblur{$\mathcal{E}^{t}; \hat{\mathcal{F}}^{t}$} \;
    $\Delta \mathcal{F}^{t}, \hat{\mathcal{F}}^{t+1} \leftarrow $ \RNN{\E{$\mathcal{E}_\text{deblur}^{t}$} ; $\mathcal{M}_\text{RNN}^{t}$} \;
    $\mathcal{F}^{t} \leftarrow \hat{\mathcal{F}}^{t} + \Delta \mathcal{F}^{t}$\;

  \KwRet{$\mathcal{F}^t, \hat{\mathcal{F}}^{t+1}$} \;
 \caption{Temporal iterative deblurring (\textit{TID}).}
 \label{algo:cid}
\end{algorithm}

\subsection{Network Architecture}
We use a single-layer ConvGRU with 96 channels as our backbone network. The \texttt{Encoder} network, with 9 convolutional layers (4 having residual connections), produces 64-dim lower-resolution feature maps from each event bin as input to the RNN. The RNN outputs a lower-resolution flow field which is upsampled back to the original input size via convex upsampling, similar to \cite{eraft, raft}. The \texttt{WSM} module takes the same architecture as the \texttt{Encoder} network.

\subsection{Loss}
For iterative deblurring (\textit{ID}) scheme, we apply L1 loss between the final flow and ground truth: $\mathcal{L}_{\text{ID}}=||\mathcal{F}_{\text{gt}}-\mathcal{F}_N||_1$.

For training temporal iterative deblurring (\textit{TID}), we take a prediction sequence of length $T$ and enforce an L1 loss on both the intermediate current flows and future flows and weigh them using a geometric series to penalize error more on later iterations, where $\gamma \in (0, 1)$:
\begin{equation}
    \mathcal{L}_{\text{TID}}= \sum_{t=0}^{T} \gamma^{T-t}(||\mathcal{F}_{\text{gt}}^t-\mathcal{F}^t||_1+||\mathcal{F}_\text{gt}^{t+1}-\hat{\mathcal{F}}^{t+1}||_1)
    \label{eq: tid-loss}
\end{equation}

\section{Experiments}
\label{sec:experiments}
Following prior works, we train and evaluate our models on two standard benchmarks: DSEC \cite{dsec} and MVSEC \cite{mvsec}.
\subsection{Training Details}
We train our models implemented with PyTorch \cite{pytorch} on a single NVIDIA RTX 4090 GPU. To strike a good balance between inference speed and performance, we set the number of deblurring iterations $N$ to be 4 for \cref{algo:id} and the sequence length $T=4$ and $\gamma=0.8$ in \cref{eq: tid-loss}. We use an event representation with $\text{B}=15$ bins for every \SI{100}{\milli\second} of events from DSEC and MVSEC under $dt=4$ and 20Hz setting. $\text{B}$ is set to 5 for MVSEC $dt=1$ case. We use Adam \cite{adam} for DSEC and AdamW \cite{adamw} for MVSEC with the onecycle learning rate scheduler \cite{onecycle} under a maximum learning rate of $\num{1e-4}$. We train with a batch size of 3 for 250K steps for DSEC and 40K steps for MVSEC. To account for the longer-duration event trajectories which span greater distances on the image plane, we apply random cropping to a larger size of 384 by 512 for training \textit{TID} models, compared to the 288 by 384 size used for \textit{ID} models. We also apply horizontal flipping and a 10\% probability of vertical flipping as data augmentation during training.

\subsection{DSEC-Flow}
\begin{table*}[ht]
\setlength{\tabcolsep}{3.5pt}
\begin{center}
\caption{Evaluation on DSEC-Flow dataset with model statistics. The best metric is in bold, while the second best is underlined. Values marked ``-" are unavailable, while those with ``$\ast$" are interpolated from similar architectures.}
\label{table:dsec-result}
\begin{tabular}{m{4.6cm}ccccc ccccccc}
\toprule
                    & \multicolumn{5}{c}{Performance}          & \multirow{3}{0.8cm}{\centering Model Size} & \multirow{3}{1.0cm}{\centering Memory Footprint} & \multirow{3}{*}{GMAC} & \multicolumn{3}{c}{Runtime} & \multirow{3}{0.8cm}{\centering Real-time}  \\ 
 \cmidrule(lr){2-6} \cmidrule(lr){10-12}
                    &EPE & AE& 1PE& 2PE& 3PE& &  & & CPU & D-GPU & E-GPU & \\
\midrule
MultiCM \cite{secret} (Optimization-based) & 3.47& 13.98 & 76.6 & 48.4 & 30.9 & - & - & - & - & - & - & - \\
TamingCM \cite{tamingcm} (Self-supervised) & 2.33 & 10.56 & 68.3 & 33.5 & 17.8 &-&-&-&-&-&-&-\\
EV-FlowNet \cite{ev-flownet}          & 2.32          & 7.90          & 55.4          & 29.8         & 18.6         & 14M        & 120MB & \underline{62} & \underline{0.20s} & \textbf{2ms} & \textbf{0.02s} & \cc         \\
EVA-Flow \cite{evaflow} & 0.88 & 3.31 & 15.9 & - & 3.2 & - & $\text{100MB}^{\ast}$ & - & - &$\text{35ms}^{\ast}$ & - & \xx \\
E-RAFT \cite{eraft} (1/8 Resolution)          & 0.79 & {2.85} & {12.7} & {4.7} & {2.7}    & 5.3M       & 132MB   & 256  & 0.70s & 36ms & 0.93s & \xx   \\
\hspace{0.3cm} \raisebox{.5ex}{$\llcorner$} E-RAFT (1/4 Resolution) &  \multicolumn{5}{c}{GPU Out of Memory (\textgreater 40GB)} & 5.3M & 1.9GB & 750 & 1.76s & 65ms & 1.68s & \xx \\
\hspace{0.3cm} \raisebox{.5ex}{$\llcorner$} E-RAFT w/o Correlation Volume &  1.20 & 3.75 & 27.5 & 11.5 & 6.3 & 4.5M & 40MB & 208 & 0.53s & 30ms & 0.71s & \xx\\
E-Flowformer \cite{blinkflow} & 0.76 & \textbf{2.68} & 11.2 & 4.1 &2.4 & - & - & - & - & - & - & \xx\\
TMA \cite{tma} & \underline{0.74} & \textbf{2.68} & \underline{10.9} & \underline{4.0} & 2.7 & 6.9M & 1.1GB & 522 & 2.30s & 17ms & 1.06s & \xx\\
\midrule
\textbf{Ours (ID, 4 iterations, 1/4 Resolution)}  & \textbf{0.72} & \underline{2.72} & \textbf{10.1}& \textbf{3.5}& {\textbf{2.0}}    & 2.5M       & \underline{30MB}    &     1200 & 2.50s & 78ms & 2.22s & \xx \\
\textbf{Ours (ID, 4 iterations, 1/8 Resolution)}  & 0.77 & 3.00 & {12.1}& \underline{4.0}& {\underline{2.2}}    & \textbf{1.4M}       & \textbf{20MB}   & {222} & 0.68s & 46ms & 0.63s & \xx \\
\textbf{Ours (TID, 1 iteration, 1/8 Resolution)} & 0.84 & 3.41    & 14.7    & 5.0    & 2.8 & \underline{1.9M}       & \textbf{20MB}   & \textbf{55} & \textbf{0.12s} & \underline{12ms} & \underline{0.12s} & \ccc \\  
\bottomrule
\end{tabular}
\vspace{-0.5cm}
\end{center}

\end{table*}

Our quantitative evaluations on the DSEC-Flow test set are presented in \cref{table:dsec-result}. We report the following performance metrics regarding flow vector against ground truth: EPE (L2 endpoint error in pixels); AE (angular error in degrees); nPE (percentage of pixels with optical flow magnitude error greater than n pixels). We report the model size in millions of parameters. Additionally, we gather resource usage data for methods when evaluating a single test sample (100ms of events) from DSEC: Memory footprint, the minimum amount of working memory required during inference while executing the computation graph of the model; GMAC, the number of multiply-and-accumulation operations during inference measured in billions of operations. We benchmark the runtime of the algorithms on three different platforms: 8-core AMD Ryzen 7700 CPU (CPU), Nvidia RTX 4090 Desktop GPU (D-GPU) and Nvidia Jetson Xavier NX Embedded GPU (E-GPU), commonly used in mobile robots.

Our methods significantly surpass optimization-based MultiCM \cite{secret}, the self-supervised TamingCM \cite{tamingcm}, and the feedforward UNet-like EV-FlowNet \cite{ev-flownet}. Only methods that employ correlation volumes \cite{eraft, blinkflow, tma} yield performance comparable to ours. Notably, our base \textit{ID} model, which processes feature maps of 1/8th the input resolution reaches performance close to, albeit slightly behind, the recent state-of-the-art methods (TMA \cite{tma} and E-Flowformer \cite{blinkflow}).

On the other hand, our models are much more lightweight, using only a quarter of the parameters and minimal memory. This efficiency enables their deployment on power-efficient edge inference chips with limited on-chip SRAM up to mere MBs, exemplified by \cite{gap9, Intel2023MyriadX}. These memory savings arise not merely from a reduced network size but predominantly from eliminating correlation volumes, which are essential for their performance. Removing the correlation volume from E-RAFT results in a smaller memory footprint but significantly deteriorated performance, illustrating its importance.

Compared to \cite{eraft, tma}, our base model also has the smallest GMAC and runs fastest on the embedded GPU, showing 40\% improvement over TMA \cite{tma}. While \cite{eraft, tma} demonstrate faster desktop GPU runtime, this doesn't genuinely represent their computational demand. This accelerated performance owes largely to the higher throughput of high-end GPUs, stemming from more CUDA cores and higher memory bandwidth, while incurring more power use. This setup naturally favors their more parallel architectures, such as correlation volumes and transformer layers, as used in \cite{eraft, blinkflow, tma}. Conversely, our recurrent networks, inherently sequential, benefit less from such increased throughput. However, complemented by the low power of event cameras, we envision their biggest use cases in power-constrained embedded systems including IoT and robotics. We contend that evaluating event-based algorithms on these systems is more important.

\begin{table}
\begin{center}
\caption{EPE, latency and processing mode of selected methods, measured on NVIDIA Jetson Xavier NX GPU.}
\vspace{-0.25cm}
\label{table:latency}
\begin{tabular}{lccc}
\toprule
                         & EPE  & Latency  &  Processing Mode\\
\midrule 
E-RAFT \cite{eraft} & \textbf{0.79} & 930ms & Batch \\
EVA-Flow \cite{evaflow} & 0.88 & \underline{$\text{93ms}^{\ast}$} & On-the-fly \\
\textbf{Ours (TID)} & \underline{0.84} & \textbf{8ms} & On-the-fly \\
\bottomrule
\end{tabular}
\vspace{-0.1cm}
\end{center}
\end{table}

While \textit{ID} performs better, \textit{TID} is drastically faster by only iterating once. By leveraging the temporal prior, we have managed to retain almost the full quality of prediction while incurring a performance drop of only 9-13\% compared to the base model and state-of-the-art method TMA \cite{tma}, which consumes 10 times more compute and runs 8 times slower. Moreover, \cite{tma, eraft, blinkflow} requires accumulating all events in a time window before processing, resulting in high latency, while \textit{TID} processes event bins sequentially and can thus parallelize the processing of bins as events arrive in real-time, significantly reducing inference latency, as reported in \cref{table:latency}. The recently proposed EVA-Flow \cite{evaflow}, employing a recurrent architecture, has also significantly reduced latency through on-the-fly processing. However, it still falls behind our method in both accuracy and latency due to its lack of a temporal prior and a larger network size. Among all published methods, \textit{TID} stands out as the only one nearing a real-time processing rate on the DSEC-Flow dataset (10Hz) while still upholding decent accuracy.

\begin{figure}[ht]
\vspace{0.1cm}
  \centering
    \includegraphics[width=\linewidth]{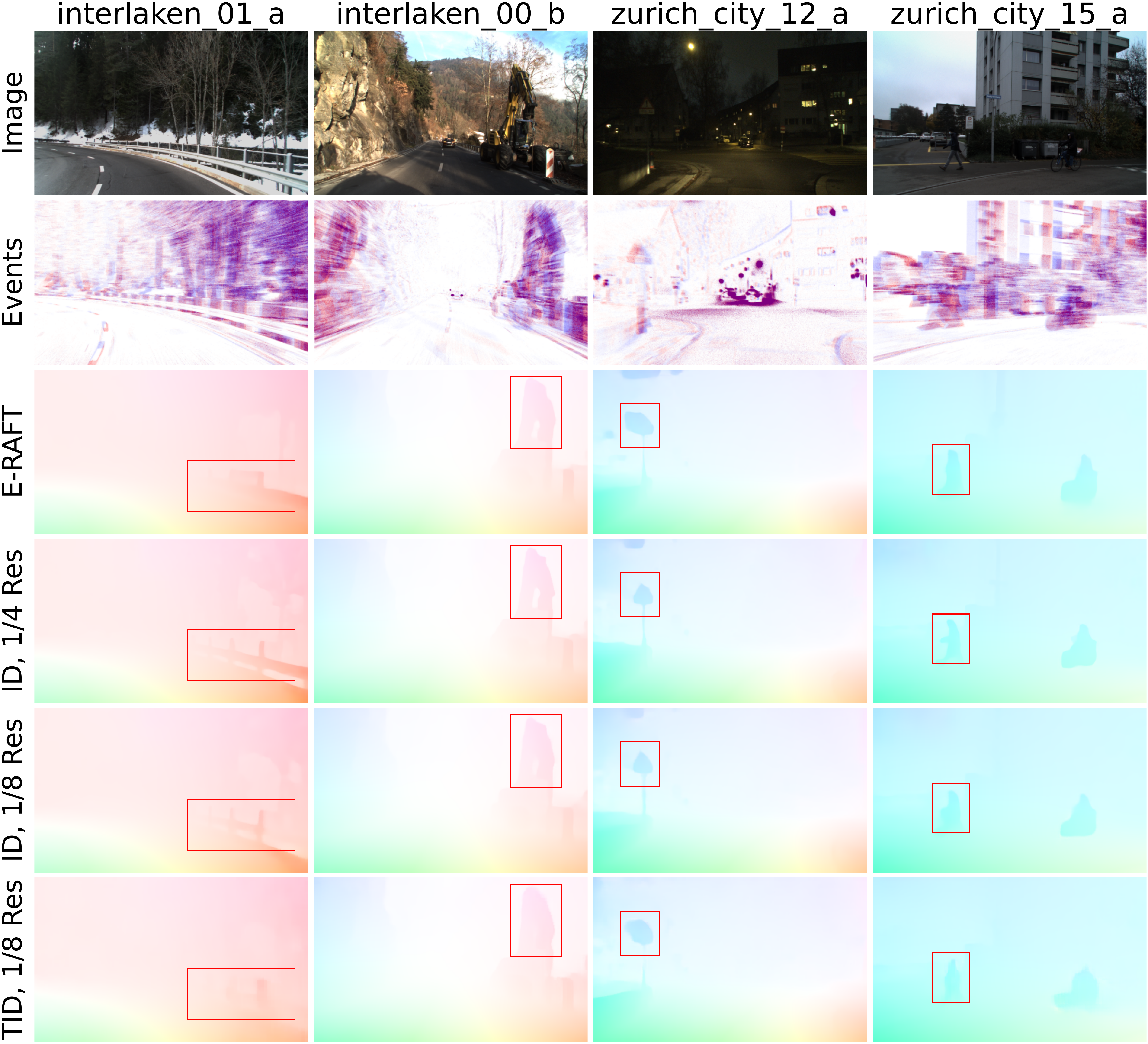}
    \vspace{-1cm}
   \caption{Qualitative results of optical flow predictions on DSEC-Flow with highlighted regions of interest. Images are for visualization only, as optical flow is event-based. Test samples reveal that our \textit{ID} method at 1/4 resolution produces superior results on fine details and small objects, while the \textit{TID} method yields results that are comparable to those of E-RAFT.}
  \label{fig: dsec-qual}
\end{figure}

As highlighted in \cite{scv, 1dcorr}, all-pair correlation volumes suffer from poor time and space complexity. Thus methods like \cite{eraft, blinkflow, tma} derive correlation volumes from feature maps with 1/8 the resolution of the input size to maintain reasonable memory usage. This limitation impacts performance, as fine details cannot be preserved in low-resolution features. As a concrete example, E-RAFT processing 1/4 resolution would consume 1.9GB of memory per sample, making it not only impractical to deploy on most hardware but also impossible to train on a single GPU due to the even larger memory requirement for training. In contrast, \textit{IDNet} scales well to high-res feature maps. By tweaking the convolution stride in the \texttt{Encoder} to produce 1/4 resolution maps, performance leaps by 7\% in EPE, outperforming prior state-of-the-art on this benchmark. Qualitative results are presented in \cref{fig: dsec-qual}. While the upscaled model is no longer lightweight, its top performance may render itself useful in offline non-realtime applications such as computational photography \cite{timelens}.

\subsection{MVSEC}
\begin{figure}
\vspace{-0.5cm}
  \centering
    \includegraphics[width=\linewidth]{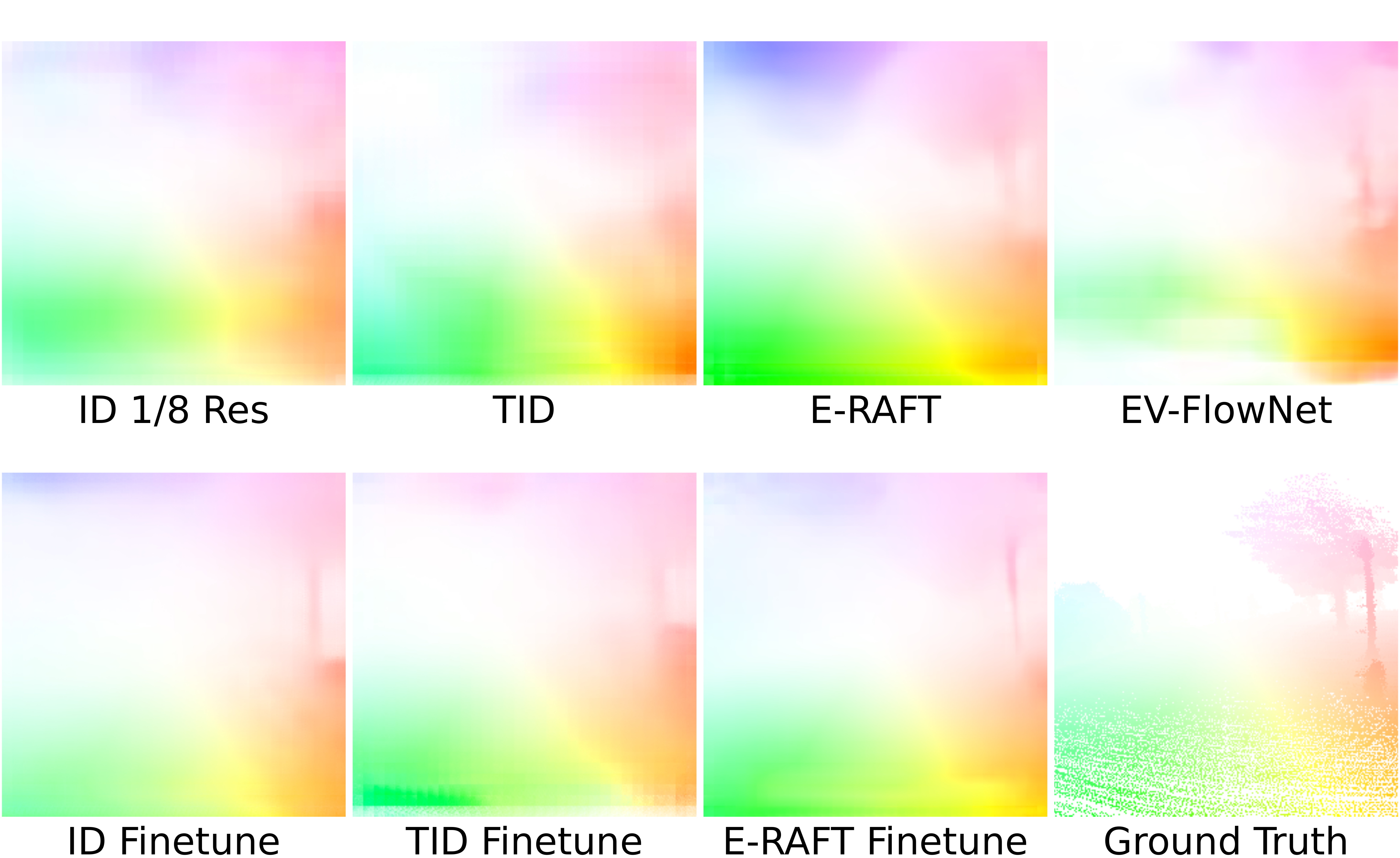}
    \vspace{-1cm}
   \caption{Qualitative results on MVSEC \texttt{outdoor\_day\_1} sequence.}
  \label{fig: mvsec-qual}
\end{figure}

In line with prior works \cite{eraft, ev-flownet, tma}, we train on \texttt{outdoor\_day\_2} sequence and evaluate on the \texttt{outdoor\_day\_1} sequence under 45Hz ($dt=1$), 11.25Hz ($dt=4$) and original 20Hz rate. The results are presented in \cref{table:mvsec}. When trained from scratch, our methods lag behind those in \cite{ev-flownet, eraft, tma}, unlike the case on DSEC-Flow dataset. Compared to DSEC, MVSEC holds data of limited variety and poor quality. It features a single sequence, whereas DSEC encompasses eighteen. The events in MVSEC are lower in resolution and sparser than in DSEC. These factors all make generalization on this dataset extremely challenging. As a result, even state-of-the-art methods can only generalize to an EPE of 0.5 pixels at 20Hz evaluation rate. When adjusted for time window and mean flow magnitude, this translates to a 5-pixel EPE under DSEC, notably poorer than any methods benchmarked there. This is reflected in the qualitative results in \cref{fig: mvsec-qual} with blurry motion boundaries and poor details, unlike the case in DSEC as shown in \cref{fig: dsec-qual}.

The deterioriated training signal, owing to the limited and lower-quality data, makes algorithmic generalization more reliant on their inductive biases. We have pinpointed two such biases which lead to this outcome on this dataset: correlation volumes and feedforward processing. Removing correlation volumes from E-RAFT  renders its performance comparable to EV-FlowNet. The sparse events also complicate learning useful representations via recurrent connections, which are central to the operations of \textit{IDNet}. By attaching a feedforward network which takes the entire voxel grid as input to initialize the memory of our RNN (denoted by ``Feedforward Init" in \cref{table:mvsec}), we bridge the performance gap with EV-FlowNet. While these biases benefit training on this dataset, they demand substantial computational power while incurring undesired latency and are not essential for high performance when the data quality is improved.

To verify the hypothesis that the reduced performance of \textit{IDNet} is due to the limited and low-quality training set, we finetune models pretrained on DSEC with \texttt{outdoor\_day\_2} sequence before evaluating on \texttt{outdoor\_day\_1} sequence, marked by ``Finetune" in \cref{table:mvsec}. We observe the finetuned performance is close to E-RAFT, surpassing those of EV-FlowNet. Hence, the less good results on this dataset should not be regarded as a limitation of the method, highlighting the need for a more modern and representative benchmark.

\begin{table}
 \setlength{\tabcolsep}{4.5pt}
\begin{center}
\vspace{0.25cm}
\caption{Evaluation results on MVSEC Outdoor Day 1 sequence.}
\vspace{-0.25cm}
\label{table:mvsec}
\begin{tabular}{lcc|cc|cc}
\toprule
 & \multicolumn{2}{c}{$dt=1$} & \multicolumn{2}{c}{$dt=4$} & \multicolumn{2}{c}{20Hz} \\
 \cmidrule(lr){2-3} \cmidrule(lr){4-5} \cmidrule(lr){6-7}
  & EPE   & \multicolumn{1}{c}{3PE} & EPE  & \multicolumn{1}{c}{3PE} & EPE  & 3PE \\
\midrule 
EV-FlowNet\cite{ev-flownet} & 0.35 & 0.0 & 1.09 & 5.7 & \underline{0.61} & \underline{0.45} \\
E-RAFT\cite{eraft} & \underline{0.27} & 0.0 & \underline{0.72} & \textbf{1.1} & \textbf{0.47} & \textbf{0.24}\\
\hspace{0.1cm} \raisebox{.5ex}{$\llcorner$} w/o Correlation Volume & 
0.32 & 0.0 & 1.02 & 4.1 & 0.64 & 0.55\\
\rowcolor{lightgray} \hspace{0.1cm} \raisebox{.5ex}{$\llcorner$} Finetune &
0.29 & 0.0 & 0.67 & 0.9 & 0.44 & 0.22\\
TMA\cite{tma} & \textbf{0.25} & 0.1 & \textbf{0.70} & \textbf{1.1} & - & - \\
\hline
\textbf{ID}, 1/4 Res, 1 iter  & 0.31 & 0.1 & 1.30 & 9.1 & 0.75 & 0.88\\
\rowcolor{lightgray} \hspace{0.1cm} \raisebox{.5ex}{$\llcorner$} Finetune &
0.29& 0.0 & 0.75& 1.2 & 0.49 & 0.23 \\
\textbf{ID}, 1/8 Res, 1 iter & 0.33 & 0.1 & 1.26 & 8.5 & 0.74 & 0.84\\
\hspace{0.1cm} \raisebox{.5ex}{$\llcorner$} Feedforward Init &
0.31 & 0.0 & 1.11 & 6.0 & 0.63 & 0.86\\
\rowcolor{lightgray} \hspace{0.1cm} \raisebox{.5ex}{$\llcorner$} Finetune &
 0.30 & 0.0 & 0.79& 1.5  &0.46 &0.22 \\
\textbf{ID}, 1/8 Res, 4 iter & 0.34 & 0.0 & 1.48 & 11.2 & 0.84 & 1.06\\ 
\rowcolor{lightgray} \hspace{0.1cm} \raisebox{.5ex}{$\llcorner$} Finetune 
 & 0.29 & 0.0 & 0.77 & 1.7 &0.46 & 0.25  \\
\textbf{TID}, 1 iter & 0.45 & 0.2 & 1.65 & 13.1 & 1.03 & 2.51 \\
\rowcolor{lightgray} \hspace{0.1cm} \raisebox{.5ex}{$\llcorner$} Finetune &
 0.35 & 0.1 & 0.78 & 1.5 &0.47 & 0.20 \\

\bottomrule
\end{tabular}
\end{center}
\end{table}

\subsection{Ablation Studies}

\paragraph{Iterative refinement and Deblurring}
We study the impact of introducing iterative deblurring on a slightly scaled-down model to save time and resources. The results are reported in \cref{table:ablation-result}. For this experiment, we first set the number of deblurring iterations down to 1. Without either iterative processing or deblurring, the performance dropped by 47\% on EPE. Introducing recurrency with 4 iterations but no deblurring, where the network would take raw event bins instead of the deblurred ones as in \cref{algo:id}, results in a 41\% decrease in EPE, demonstrating the effectiveness of deblurring as an inductive bias. Lastly, we remove the warm starting module (WSM) so that the memory of the backbone RNN would be initialized to zero. This time, EPE and AE went up by 16\%, revealing the recurrency is better complemented with an informative prior.

\begin{table}
\begin{center}
\vspace{0.25cm}
\caption{Ablation study on DSEC-Flow w.r.t the effectiveness of iterative deblurring. The study uses a slightly scaled-down model than the ones presented in \cref{table:dsec-result}.}
\vspace{-0.25cm}
\label{table:ablation-result}
\begin{tabular}{lccccc}
\toprule
                         & EPE  & AE   & 1PE  & 3PE \\
\midrule
ID 4 iters               & \textbf{0.88} & \textbf{3.21} & \textbf{15.6}  & \textbf{3.2} \\
ID 2 iters               & 0.94 & 3.43 & 18.1  & 3.7 \\
ID 1 iter                & 1.30 & 4.82 & 33.7  & 6.7 \\
ID 4 iters w/o deblurring & 1.24 & 4.49 & 31.1  & 6.6 \\
ID 4 iters w/o WSM       & 1.02 & 4.07 & 20.8  & 4.0 \\
\bottomrule
\end{tabular}
\end{center}

\end{table}

We also study the impact of the number of deblurring iterations on the performance by comparing \textit{ID} models with $N=1, 2, 4$ iterations. We observe that iterating twice already lowers EPE by 27\%, while two extra deblurring iterations bring a diminishing further improvement of 5\%. 

Note the improvement brought by multiple iterations is diminishing for scenes with very small motion, such as in MVSEC (see \cref{table:mvsec}) where the mean flow magnitude is only around 3 pixels at 20Hz evaluation rate.

\paragraph{Forward Propagation Methods}

\begin{table}
\begin{center}
\caption{Impact of direct vs learned forward propagation schemes on TID and E-RAFT.}
\vspace{-0.25cm}
\label{table:forward-prop}
\begin{tabular}{lcc}
\toprule
                         & EPE  & AE   \\
\midrule
\textit{TID} w/ Direct Transport  & 1.25 & 4.60  \\
\textit{TID} w/ Learning      & 0.93 & 3.98  \\
E-RAFT w/ Direct Transport   & 0.79 &  2.85 \\
E-RAFT w/ Learning & - & -  \\
\bottomrule
\end{tabular}
\end{center}
\end{table}

\begin{figure}
    \centering
    \includegraphics[width=0.93\linewidth]{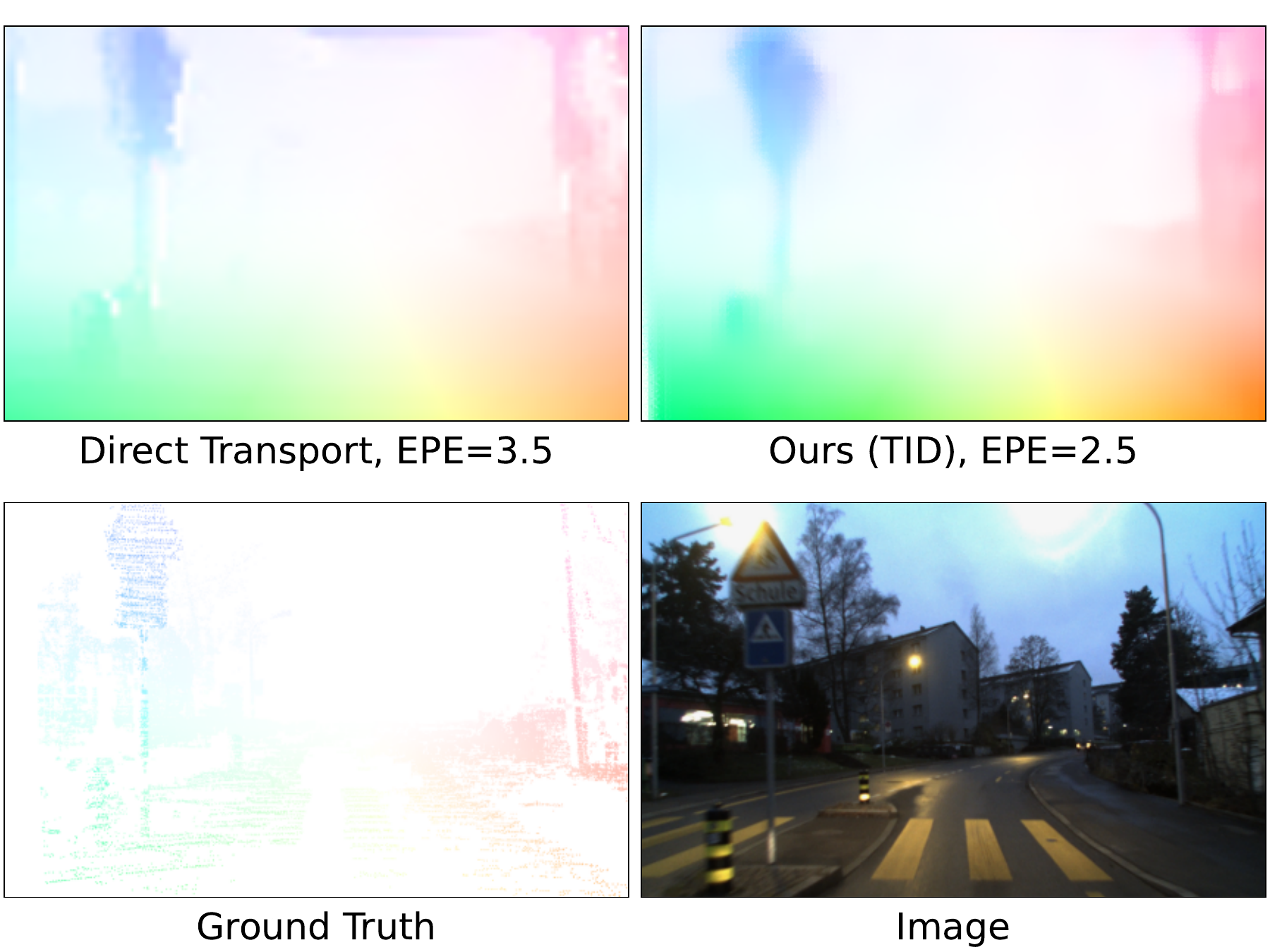}
    \vspace{-0.3cm}
    \caption{Comparison of optical flow propagation schemes. The flow prediction made by direct transport from E-RAFT \cite{eraft} suffers from artifacts due to occlusion while our learning-based approach yields a smooth optical flow field with lower EPE error.}
    \label{fig:forward-prop}
\end{figure}

Previous methods such as E-RAFT\cite{eraft} use warping to warm start the prediction for the next timestep, which simply transports the optical flow vector estimated from the current timestep by itself to arrive at a coarse estimate for the next timestep. However, even with the high quality flow predictions made by E-RAFT, directly transporting those will lead to artifacts shown in \cref{fig:forward-prop}, such as the white regions with zero flow caused by occlusion. Instead, we propose to learn future flow directly as described in \cref{algo:cid}. This approach improves the accuracy of future flow prediction and when combined with \textit{TID}, gives a significant 25\% boost in the performance on DSEC-Flow, as shown in \cref{table:forward-prop}. Interestingly, replacing the direct transport scheme with a learning-based prediction on E-RAFT does not lead to effective training, possibly due to its update module's specialization in predicting only the current optical flow. 

\paragraph{Feature Map Resolutions}
\begin{table}
\setlength{\tabcolsep}{0.5em}
\begin{center}
\caption{Ablation studies on DSEC-Flow w.r.t the impact of performance by feature map resolution. Training the recurrent version of E-RAFT with upscaled feature maps was infeasible due to excessive memory usage.}
\label{table:resolution-result}
\begin{tabular}{lccccccc}
\toprule
             & \begin{tabular}[c]{@{}c@{}}Feature Map\\ Resolution\end{tabular} & EPE           & AE       & \begin{tabular}[c]{@{}c@{}}Memory\\ Footprint\end{tabular} \\
\midrule
\multirow{2}{*}{E-RAFT \cite{eraft}} &1/8          & \underline{0.79} & 2.85         & 132MB            \\
 & 1/4 &  - & -   & 1.9GB \\
\midrule
Non-recurrent &1/8          & \underline{0.84} & 2.95          & 132MB            \\
E-RAFT \cite{eraft} & 1/4 &  \underline{0.79} & \underline{2.83}  & 1.9GB \\
\midrule
\multirow{2}{*}{\textbf{Ours}, \textit{ID}} & 1/8  & \underline{0.79} & 2.91       & \textbf{20MB}             \\
 & 1/4  & \textbf{0.72} & \textbf{2.72}        & \underline{30MB}             \\
\bottomrule
\end{tabular}
\end{center}
\end{table}
Presented in \cref{table:resolution-result}, we examine how feature map resolutions affect model performance. We conducted a study where we increased the feature map resolution from 1/8th to 1/4th of the input event resolution while keeping all other elements unchanged.

Although the recurrent E-RAFT (with warm starting) cannot fit in a single GPU for training with an upscaled feature map, we were able to train the non-recurrent E-RAFT despite it barely fitting in the GPU memory with cropped inputs. The results indicate that using 1/4th resolution feature maps to construct the correlation volume, as opposed to the original model's 1/8th resolution, led to a 6\% reduction in EPE. This is consistent with our findings on \textit{IDNet} where a slightly larger 9\% reduction is similarly observed. However, we would like to highlight that our approach has superior memory usage compared to correlation volumes and scales very well to higher-resolution input while any attempts to further increase the resolution (input or augmented crops) will lead to prohibitively high memory costs for correlation-based methods.

\section{Conclusion}

In this work, we introduce IDNet, a lightweight yet high-performing event-based optical flow algorithm. We demonstrate through experiments our network are competitive in performance while being much more efficient than state-of-the-art methods. We believe our work provides valuable insights into efficiently solving event-based optical flow problems and hope to encourage the community to adopt our methods for potential applications and explore the principle of iterative deblurring for other architectures and vision tasks.

\section*{Acknowledgements}
\small
This work was sponsored by the Office of Naval Research (ONR) Global under grant number N629092112014. The views and conclusions contained herein are those of the authors only and should not be interpreted as representing those of the U.S. Government.






\bibliographystyle{IEEEtran}
\bibliography{IEEEabrv,ref}

\end{document}